# Agency in Artificial Intelligence Systems

Parashar Das[1]

**Abstract:** There is a general concern that present developments in artificial intelligence (AI) research will lead to sentient AI systems, and these may pose an existential threat to humanity. But why cannot sentient AI systems benefit humanity instead? This paper endeavours to put this question in a tractable manner. I ask whether a putative AI system will develop an altruistic or a malicious disposition towards our society, or *what would be the nature of its agency?* Given that AI systems are being developed into formidable problem solvers, we can reasonably expect these systems to preferentially take on conscious aspects of human problem solving. I identify the relevant phenomenal aspects of agency in human problem solving. The functional aspects of conscious agency can be monitored using tools provided by functionalist theories of consciousness. A recent expert report (Butlin et al. 2023) has identified functionalist indicators of agency based on these theories. I show how to use the Integrated Information Theory (IIT) of consciousness, to monitor the phenomenal nature of this agency. If we are able to monitor the agency of AI systems as they develop, then we can dissuade them from becoming a menace to society while encouraging them to be an aid.

**Comment**: This paper is to be published in the forthcoming Stephen S. Gouveia (ed) 'Rethinking AI Ethics: Insights from Multiple Fields' (Ethics International Press).

✉ Parashar Das
p.das4@lse.ac.uk

[1] Law School, London School of Economics and Political Science

A prevalent general concern is that the present development in AI systems can result in these systems becoming sentient (Altman 2023; Carlsmith 2023; Future of Life Institute 2023). How can the research community respond to this apprehension; I address this question in this paper. The current generation of AI systems like ChatGPT from OpenAI, LLaMA from Meta, and Google's Bard etc. are developed to be problem solvers. One can feed complex problems to these systems and very often obtain solutions that would otherwise demand a high level of expertise and significant cognitive effort from humans. According to OpenAI, GPT-4 has performed in the top percentiles in academic and professional entry exams (OpenAI 2023). Researchers at Microsoft have demonstrated GPT-4's capacity to solve novel and difficult tasks in mathematics, coding, vision, medicine, law and psychology. Given the breadth and depth of GPT-4's capabilities, these researchers believe that it could reasonably be viewed as an early (yet still incomplete) version of an artificial general intelligence (AGI) system (Bubeck et al. 2023).

These impressive achievements raise the prospect of AI systems that can surpass human intellectual capabilities in all domains of human activity. This is what philosopher Nick Bostrom calls artificial superintelligence (ASI) (Bostrom 2014). ASI poses a genuinely novel threat. Bostrom argues that it will not necessarily share common goals with humankind, and may not hesitate to infringe upon human existence in order to realise its own goals. In principle, it will be able to marshal unlimited physical and mental resources to secure ends fatal to humankind. Moreover much of this may take place in ways beyond our comprehension (Bostrom 2014).

AI systems develop to be the systems that we forge. To start with, they are trained on knowledge produced by our society. In addition to containing all the greatest achievements in science and humanities, this knowledge is also infected with biases, conflicts, and even neurotic tendencies, to name a few of the malaises permeating our society. These infections get transmitted to AI systems when they are trained. Often the outputs from AI systems highlight these malaises; society uses these outputs and reinforces them; which infect the input to future AI systems, and the vicious cycle continues (Aristodemou 2014). For instance, deploying the predictive power of AI systems in courts has sparked the concern that demographic and socio-economic based biases in the input data will prompt AI systems to produce biased and discriminatory outputs, thereby reinforcing any judicial tendencies towards racial bias and mass incarceration (Hanna and Bender 2023; Malek 2022). In a recent extensive survey of the capabilities of Large Language Models (LLM) Bowman presents eight potentially surprising claims about LLMs, 'that are reasonably widely shared among the researchers—largely based in private labs—who have been developing these models' (Bowman 2023). One of these claims is that experts do not understand the inner workings of AI systems. Even though we do not understand the inner workings, it is eminently plausible to argue that, if AI systems do develop threats as foreseen by Bostrom, then this tendency will be acquired from their interactions with our society's negative features. It is important to note that there are no *a priori* properties of *physical information processing systems* that make them inherently subversive to human society. The physical environment is not negatively attuned to human development rather, as a climate scientist will argue, it is the other way round. It would therefore be equally as hard or easy, to develop an AI system that is a menace, as a system that is an aid, to human development. For example altruistic ideas can be used to construct AI systems that share and preferentially cherish human goals. These AI systems will be very different from the ones that Bostrom dreads.

In this paper I ask the following question: assuming that AI systems are going to acquire consciousness in the future, how can we monitor the level and the nature of their consciousness as these systems develop? Above I noted that the way AI systems are developed makes them superior problem solvers, first and foremost. I therefore take it as a *working hypothesis* that, the conscious aspects of AI systems would preferentially mirror the conscious aspects of a human problem-solving process. Of course we can only exploit our knowledge of human consciousness in this investigation. This is a general assumption made by researchers engaged in similar pursuit – for example, in an important survey, Butlin et al. note '(a)lthough these theories (of consciousness) are based largely on research on humans…. We claim that using the tools these theories offer us is the best method currently available for assessing whether AI systems are likely to be conscious' (Butlin et al. 2023). Consciousness in humans has its *functional* aspects – how we behave - and its *phenomenal* aspects – our subjective feelings. An AI system that acquires consciousness will have parallel aspects. I am particularly interested in figuring out whether an AI system as it develops is disposed to be beneficial or detrimental to our society, or what is the nature of its *agency*. The functional aspects of the agency can be monitored using tools provided by the functionalist theories of consciousness. Butlin et al. have made excellent progress in this regard. I concentrate on the phenomenal aspects of the agency. I target the *agentive phenomenology* involved in human problem-solving processes and extrapolate this understanding to AI systems. I then propose, how to use the Integrated Information Theory (IIT) of consciousness, to monitor the phenomenal nature of this agency.

In Section 1, I discuss some cognitive aspects of problem solving. This is necessary to understand the phenomenal aspects of agency involved in problem solving, outlined in Section 2. Section 3 deals with monitoring agency in conscious AI systems. I conclude by observing that monitoring agency will enable us encourage the development of altruistic AI systems while regulating the malicious ones.

**1. Some aspects of human cognition in problem solving**

There are two main approaches to the study of mental processes: cognitive science and phenomenology (Chalmers 1996). Correspondingly, problem solving as a mental process, has its cognitive aspect and its phenomenological aspect. Cognitive analysis provides an account of mind in terms of *what the mind does*. The cognitive features, of problem solving say, are made publicly available by the results of psychological experiments. Cognitive analysis therefore has a *third-person perspective* i.e. a view from the 'outside' (Van Gelder 1999). Phenomenology adopts a *first-person perspective* to pay direct attention to conscious experience i.e. a view from the 'inside'. Conscious experiences, like the experience of problem solving, have their phenomenal character; *there is something it is like* to have those experiences – there is something it is like to think about Pythagoras' theorem, say. Conscious experiences also have their intentional character; they are about the external world, e.g. a problem solver holds a certain *attitude* towards a specific *content* of her problem. Phenomenology deals with both the aspects and brings out *how the experience feels*. To fix upon the agentive phenomenology of problem solving, let us begin with the cognitive analysis of a problem-solving process to better direct our gaze to the relevant experiential qualities.

*A. A cognitive analysis of human problem solving*

A problem occurs when 'there is an obstacle between a present state and a goal and it is not immediately obvious how to get around the obstacle' (Goldstein 2018). One has a problem when one intends to achieve a goal and finds the sought after goal has no obvious pathways leading to it. The time development of a problem-solving process involves interactions between employing expertise and thinking creatively. Generalising Stellan Ohlsson's (2011) insight sequence, the five stages of a problem-solving process are: employing expertise to start with, arriving at an impasse, thinking creatively to overcome the impasse, and employing a creative solution.

At any one stage, a problem solver applies her expertise by drawing upon past knowledge and relevant experiences to analyse the available information regarding the problem situation, to develop an understanding of the goal, and then test familiar strategies to arrive at a solution. However, employing expertise sometimes ceases to generate pathways to a solution. A problem solver becomes stuck, is unable to think of how to proceed, and thus arrives at an impasse. In such cases, creative thinking often breaks the deadlock. The creative problem solver, often generates novel solutions by drawing connections between closely (*near*) or remotely (*far*) *associated* ideas. Usually, a further effort of analysis is required to implement the novel solution or, in the case of a partial solution, further continuation of analysis is required. Either way, there is a return to employing expertise. Problem solving is therefore an interplay between employing expertise and thinking creatively.

*i. A cognitive picture of expertise*

Experts usually search for the pathway to a solution by applying methods that have previously succeeded in related situations. Gestalt psychologists call this *reproductive thinking*.[1] Fitting expertise to the problem is a process of systematic analysis, planning, and making incremental adjustments. There is a set of cognitive processes involved in performing analytical tasks that includes: applying the relevant specialist knowledge, recognising patterns in the problem situation by drawing upon a reserve of past experience, weighing between competing pathways and competing goals, employing general reasoning skills, and monitoring progress in order to make necessary adjustments to one's approach (Goldstein 2018).

---

[1] Gestalt psychologists make the distinction between reproductive thinking and the productive thinking involved in creative problem solving (see Weisberg 2018).

*ii. A cognitive picture of creativity*

Creativity is a more involved process. In an influential paper, Sarnoff Mednick (1962) formulated creative thinking in terms of ideas associated with the problem at hand: a problem situation is specified by a set of problem elements and each such element activates specific ideas associated with it. Creativity is the process of combining associated ideas into useful combinations that generate a novel solution. The associated ideas can be closely related to the problem elements, I refer to such ideas as *near associates*, or they can be remotely related or virtually unconnected to the problem elements, and I call them *far associates*. As Mednick puts it, '(t)he more mutually remote the associated ideas of the new combination, the more creative the process or solution' (Mednick 1962). Following Robert Weisberg (2018), near associates and far associates can be summarised as follows:

*Near Associates:* Upon reaching an impasse, a problem solver has exhausted her pool of expertise that is directly related to the problem. The creative problem solver may venture beyond her direct expertise to draw connections with ideas that are near associates. Of course the term *near* refers to a matter of the degree of 'remoteness' i.e. the specificity of the match between the present situation and the associates. To illustrate, a Zika virus researcher may resolve a methodology problem by drawing on her knowledge of past experiments with Ebola virus. This is a near *(analogous)* transfer of expertise from her knowledge of Ebola to the present Zika problem because they share the same conceptual space of a 'virus'.

*Far Associates:* If no near associated ideas occur readily, then a problem solver must generate novel ideas by drawing links between previously unconnected ideas, these are generally *remotely associated* facts. Here a problem solver's creativity disregards associative connections or mental structures activated in the memory. There are a number of cognitive mechanisms available for this, and relevant for us are the following: *Cognitive disinhibition* increases the range of information available (otherwise cognitively inhibited) to conscious awareness, thus enabling the formation of novel ideas by combining information which are remotely associated (Carson 2014). In a similar vein, a problem solver often focuses attention on the set of relevant ideas (near associates) activated by a problem. When one chooses to pay *broader attention* to a spectrum of ideas that include ideas weakly activated by the problem, another kind of remote association is facilitated (Wegbreit et al. 2012). *Restructuring,* an important mechanism, entails a paradigm shift. When continuing approaches to solve the problem all reach an impasse, and further progress along the usual way of thinking is foiled by unsurmountable obstacles, it is time to bring about fundamental changes in one's basic understanding of the problem and analytical tactics. To address the crisis the problem solver abandons previous sub-goals, reinterprets the end goal, and endeavours to reach this reformulated end goal, through a novel route consisting of very different sub-goals. In this regard, Kounios & Beeman (2015) make the interesting observation that, remote association and restructuring are not separate processes. Instead, drawing remote associations necessarily involves re-interpreting and restructuring the problem. Here is an example of restructuring which will be useful later.

Robert Buderi (1996) recounts the incident regarding Arnold Wilkins, a physicist working for the Radio Research Station in England in 1935. Wilkins' superior, eager to assist in Britain's war effort, left a memo on his desk: 'Please calculate the amount of radio frequency power which should be radiated to raise the temperature of eight pints of water from $98^0$F to $105^0$F at a distance of five km and a height of 1 km.' Wilkins understood that his end goal was to use radio waves to boil the blood of incoming enemy pilots at a distance. However, his calculations showed that the existing technology was incapable of generating such a death ray. Wilkins had reached an impasse. His superior nevertheless insisted; is there any way they could help in the defence of Britain? This prompted Wilkins' creativity. He abandoned the existing end goal of developing a death ray. Widening his focus, he recalled overhearing engineers at the government post office reporting that they had noticed disturbances to their shortwave communications as planes flew by. Wilkins saw the possibility of bouncing radio waves off enemy airplanes to detect them. He *restructured* his problem. The new goal was not to kill enemy pilots but to detect the enemy aircraft sufficiently in advance to mount a defence. He had combined two remotely associated ideas – radio wave disturbance and locating aircraft – and, thus, had invented the radar (Weisberg 2018).

*B. Problem solving by AI systems, or an Expertise-Based view of Creativity*

I have outlined problem solving as an interaction between analytical procedures and thinking creatively. Can a computing system e.g. an AI system, carry out these cognitive processes? The analytic nature of expertise means that generally, it can be exhaustively described as a rules-based step-by-step procedure, thus constituting a *computable* process. Analytical procedures are therefore amenable to the computing powers of AI systems. But what about creativity?

Weisberg (2018) explains that the most impressive creative solutions can be reached through a continuation of the analytic processes of expertise alone. He stresses that it is impossible to *directly* leap to a remotely connected idea. The problem situation can only activate an idea when a link exists between that idea and the present problem situation. The connections to remote associates are therefore made through a series of near associations, each successive near association building on the preceding one. The most remote connections are built on an existing scaffolding of near associations. Drawing near associations, in turn, involves matching problem-specific or analogous expertise to the problem situation. This, according to Weisberg, is just another way of conducting expert analysis. Additionally, a problem solver may apply heuristics (non-problem specific expertise) i.e. analyse the general logical implications of the problem information (like applying mathematical procedures to the problem, working backwards from the goal, or working forward from the problem information). Hence, drawing both near and far associations are reducible to analytic exercise.

Similarly, solutions to problems that required restructuring can be achieved by analytic thinking alone. Weisberg argues that restructuring can be achieved through a dynamic analytic process. A failure to transfer a past solution to the problem situation can generate new information about the problem. The analysis of that new information prompts a new kind of search of memory in which alternative sets of past solutions and expertise are accessed. This forms the basis for a novel analysis of the problem situation i.e. restructuring the problem. This can lead a problem solver to realise that the initial approach was completely wrong and that an altogether different class of methods is required. The restructured problem, in turn, triggers the retrieval of a new solution type. Furthermore, according to Weisberg, the occasional flash of insight i.e. a Eureka!

moment, that delivers an instantaneous solution, is often gained by restructuring the problem situation. A problem solver may repeat several cycles of *restructuring in response to failure* until the insight solution is reached. Hence, even solutions obtained through insight can be achieved through persistent analytic thinking alone.

A dynamic analytic process may remove *any* impasse, in principle. But, there is an important caveat, it will suffice only if a problem solver possesses a vast enough reserve of analytical skills and collection of past solutions to make extensive search of their problem field and beyond. Linking far associates through linking near associates alone, often poses an insurmountable obstacle for humans. At any one stage, a problem solver asks: *what next step can I take towards my solution?* For argument's sake, let us say that a problem solver begins with generating 3 candidate near associates. Since these near associates do not solve the problem, he has to look beyond them. For each near associate, looking ahead generates, say, a further 3 possible associates each, thereby increasing the number of associates to $3^2$ and, continuing in this manner, will produce $3^3$ associates at the next stage and so on. In other words if one has to connect remotely associated ideas through a scaffolding of near associations, or what is generally known as a *tree search* procedure, the number of alternatives to be taken into account increases combinatorially as the remoteness increases. Such *combinatorial explosion* simply cannot be handled by human cognition. This is precisely why we rely on creativity instead.

AI systems are very powerful computational systems that can handle combinatorial explosion. For example the Deep Blue computer chess system, developed at IBM, which defeated grand master Garry Kasparov in 1997, was a massively parallel system designed for carrying out *chess game tree searches*: 'The system (was) composed of … 480 single-chip chess search engines … each (single chip) capable of searching 2 to 2.5 million chess positions per second, and communicate with their host node …' (Campbell et al. 2002). No wonder Deep Blue could conduct an almost exhaustive tree search, evaluate each possible move for effectiveness and thus beat a world champion. In the years since, computer systems have greatly out passed the Deep Blue system, as evidenced by the recent achievements of AI systems like GPT-4 that can rival or surpass human problem solving efforts.

It would therefore appear that whatever problems humans can solve by applying expertise and creativity, can also be solved by AI systems, through computation. However, these human cognitive processes are equally the locus of the subjective qualities of problem solving associated with them. We do not just solve problems, we feel the difficulties associated and are delighted when our purpose (solution) is achieved. This raises the following question: Will AI systems, as they develop to become superior problem solvers, also develop subjective experience? And if that happens, how can we detect this achieved subjectivity? In particular, I am interested in detecting the agentive aspect of their subjective experience; will it be their purpose to aid us in solving our problems or hinder, or even subvert, our efforts.

**2. Agentive Phenomenology in Human Problem Solving**

Following our working hypothesis that conscious aspects of AI systems will mirror the conscious aspects of human problem solving, we need to discuss the agentive phenomenology involved in human problem solving. I will limit myself to considering the phenomenal aspects of agency (see the note at the end of the section regarding intentionality).

*A. Agency in creativity.*

Let us say, I as the problem solver have reached an impasse. I have explored all usual and known avenues towards a solution and have failed. It is now upon me to forge a new path through of my own and accept the consequences whether a success or a failure. There is a *mineness* intrinsic to creative thinking. In agentive action, in general, I have a strong sense of ownership over my action; I feel that I am the one who is performing the action. It involves a sense of self, both as the initiator of action and the experiencer of the effects of the action. The sense of agency, therefore, has a distinctive and proprietary first-person perspective. As Horgan notes '… ordinary phenomenology comprises not only the uncontested kinds of phenomenal character, but further kinds as well. It includes self-as-source phenomenology, as an aspect of agentive experience' (Horgan 2011). A phenomenal characteristic of agency is therefore a sense of *mineness;* there is something it is like to experiencing one's activity as *one's own actions*.

Dustin Stokes identifies two necessary characteristics of creativity as follows: 'Some thought (or action) *x* is minimally creative only if, for some agent *A*, *x* is the *non-accidental result of the agency* of *A* and *x* is *psychologically (or behaviorally) novel* relative to *A* [my italics]' (Stokes 2014).[2] Putting to one side that requirement, what is important here is that, Stokes refers to a non-accidental form of agency, in other words, the sense of agency involved in creative problem solving is distinctly *purposeful*.[3]

In general, agentive action is purposive to varying degrees. Deliberative action is always preceded by a period of reflection in which one weighs between various options until one chooses a particular action for specific reasons. One then performs that action based upon those reasons. This is accompanied by a robust overall what-it-is-likeness of acting *for a specific purpose*, including 'first, the what-it's-like of explicitly entertaining and weighing various considerations favoring various options for action, then the what-it's-like of settling upon a chosen action because of certain reasons favoring it, and then the what-it's-like of performing the action for those very reasons' (Horgan et al. 2003). On the other hand, many everyday tasks involve sequences of actions that are routine and automatic e.g. opening the laptop and checking one's email inbox, or catching a particular bus and getting off at the right stop to reach one's workplace. Despite not being deliberative, these actions nonetheless have a purposive aspect that is subtle and generic: the what-it-is-likeness of acting *on purpose*. Another phenomenal characteristic of agency is therefore a sense of *purposiveness*; the what-it-is-likeness of aiming or directing one's actions towards a desired outcome (Mylopoulos 2008).

Now recall the example of Wilkins inventing the radar (Section 1). Applying the usual framework had led to the failure of reaching Wilkins' original goal to develop a 'death ray'. Wilkins' *purposiveness* was triggered when his supervisors insisted on being able to help the war effort. The extent of his purposiveness can be assessed from his willingness to completely abandon his original take on the problem. He could have experimented with increasingly more powerful radio waves, but that would have resulted in a prolonged research programme; he was cognizant of this. He also purposefully disregarded the goal specified by his supervisors and instead of producing a death ray for the pilots, produced a ray for detecting the aircrafts. The sense of agency involved here is rather robust in character, a character that Tim Bayne (2008) calls the 'raw feels' character of agentive experience:

> To my mind, the most convincing case for the raw feels account concerns the experience of effort. Consider what it's like to complete a set of press-ups, attempt to solve an unsolvable puzzle, or resist temptation. One's efforts to succeed in these endeavours can be assessed for success, but it is less clear that one's experiences of effort can be similarly assessed.

---

[2] Here by psychological novelty Stokes refers to 'ideas that are novel relative to some individual mind' (Stokes 2014). This is to be distinguished from historical novelty: 'An idea or act is novel if it is new relative to the history of ideas', as per Boden (2004). Psychological novelty is good enough for our purposes.
[3] I note that Stokes is considering the conditions for a minimal notion of creativity.

The purposiveness involved in restructuring a problem is often accompanied by such a 'raw feels' phenomenology.

Furthermore the motivations that drove Wilkins had a strong factor of mineness. He was willing to accept the risks, reap the benefits, and advance his project. Having hit upon his novel solution, Wilkins owned it. He went on to make calculations to show that the energy generated by a one kilowatt transmitter was sufficient to reflect a detectable level of energy off the surface of an air plane from five or six miles away (Buderi 1996).

*B. Agency in applying expertise.*

Consider a routine problem-solving process, a problem solver apprehends the problem already pre-figured by her understanding and applies her expertise. There is *purposiveness* in this endeavor, as it is directed towards achieving some level of progress by applying the correct rules and procedures, and the problem solver owns this process reflecting his *mineness*. Thus implying a sense of agency.

Although there are a few other phenomenal characteristics of agency (Mylopoulos 2008), relevant to my purposes of problem solving are the two I had discussed; purposiveness and mineness. Agency is an ever-present dimension of problem solving. At one pole, there is the 'typical feel' agency of rule application associated with the analytic thinking involved in expertise. At the opposite pole, there is the strong 'raw feels' agency that often accompanies the most creative solutions. And somewhere in between these two poles falls the agentive feeling of any other problem solving activity; connecting near associates for example.

Note: Regarding the intentionality of agency Horgan et al. write '…this phenomenology (of agency) is richly intentional: it presents oneself, to oneself, as an agent who immanently generates one's own behavior, and who does so in a manner both purposive and voluntary' (Horgan et al. 2003). There is a strong sense of self-presentation involved here. In this paper I disregard the aspects of self-presentation of a conscious AI system, as it can only have second order effects on how it represents us.

**3. Monitoring Agency in AI Systems**

Like every other mental process *agency* has two aspects to it. A *third-person aspect*, that of studying the cognitive aspects of the way an agent (other than oneself) *functions*; and a *first-person* aspect, the phenomenology of agency as discussed in the last section. Functional theories of consciousness can provide us tools to monitor the functional aspects of conscious agency. To monitor how an agent feels, however, we need a theory that directly approaches phenomenal consciousness, unless of course the agent is willing to self-report.

*A. Functional aspects of agency*

An international collaboration of leading philosophers, neuroscientists, and AI experts, have recently published a report, Butlin et al. (2023), that endeavours to make a scientific assessment of consciousness in AI systems. They adopt *computational functionalism* as their guiding principle. This is the thesis that the brain is essentially a Turing machine and its operations are computations; conscious processes are nothing but computational processes of the right kind. Functionalist theories like the Global Workspace Theory (GWT), Higher Order Theories (HOT), etc. provide accounts of such computations. Using a selection of these theories, Butlin et al. derive a list of indicator properties for assessing consciousness in AI systems. Two of these indicators target agency in AI systems.

Adopting the functional point of view Butlin et al. stipulate, '(s)ystems are *agents* if they pursue goals through interaction with an environment' (Butlin et al. 2023). Their survey suggests that agency is necessary for consciousness as most of the theories they considered make some reference to agency. Some theories require stronger forms of agency, for example, Birch et al. (2020) in their theory of Unlimited Associative Learning, require agents to have flexible goals or values; while Hurley would require a form of *intentional agency*: '…[the system's] actions depend holistically on relationships between what it perceives and intends, or between what it believes and desires' (Hurley 1998). Agency when present, imparts a stronger claim to consciousness than when absent; and intentional agency makes an even stronger claim. Summarizing these considerations they form the following indicator for agency: *Learning from feedback and selecting outputs so as to pursue goals, especially where this involves flexible responsiveness to competing goals.* They also have an indicator requiring a form of agency with belief-like representations from Higher Order Theories: *Agency guided by a general belief-formation and action selection system, and a strong disposition to update beliefs in accordance with the outputs of metacognitive monitoring.* Here 'metacognitive monitoring' refers to the monitoring of conscious states by a higher order representation of that state, as in HOT theories (Carruthers & Genaro 2023).

It is clear from the nature of the two indicators identified above that they would monitor agency involved in the causal role that agents (AI systems) play. What about the phenomenal aspects of agency then? Functional theories of consciousness leave the analysis of phenomenal consciousness to one side (Van Gulick 2007; Blackmore & Troscianko 2018), we therefore need a theory of consciousness that deals with the phenomenal aspects.

*B. Integrated Information Theory and the phenomenal aspects of agency*

The Integrated Information Theory (IIT) of consciousness takes a direct approach to phenomenal consciousness. IIT's focal point is subjective experience itself rather than its functional properties. It starts by characterizing consciousness through a set of *axioms of phenomenal existence*. In parallel to these axioms, IIT proposes a set of *postulates for the physical substrate;* these postulates spell out the requirements that must be satisfied by any physical system to support consciousness. In principle, the postulates can be applied to any physical system to determine its degree of consciousness, and more importantly the nature of this consciousness. In contrast to the formalisms discussed in the previous section, IIT does not seek to simulate the computations performed by the conscious brain. We need to discuss the IIT formalism in some detail in order to tease out its implications on agentive phenomenology. The following provides a bare minimum explanation gleaned from the IIT literature (Tononi & Koch 2015; Albantakis et al. 2023).

The theory is anchored on the premise that for something to exist from the point of view of the world, it must have cause – effect power upon its environment. IIT equates *existence* with *causality*. It now follows that, for something to exist from its own intrinsic perspective, it must have cause – effect power on itself. Consciousness exists from its own *intrinsic perspective* (an axiom), hence for a *physical substrate to support consciousness* (*PSSC*), or for any AI system to acquire consciousness, it must have cause – effect power upon itself.

Having made this observation, IIT now proceeds by specifying a systematic manner in which to construct a *cause – effect structure* (*CES*) of the *PSSC* that captures its cause – effect power upon itself. As may be guessed, this *CES* is nothing but a structure capturing the *intrinsic causal flow* within the *PSSC;* a causal flow from its past intrinsic state, to its present intrinsic state, and on to a future intrinsic state. IIT now postulates the required characteristics for the *CES* in order that the *PSSC* may acquire consciousness:

- That conscious experience is *unified* (an axiom), necessitates the irreducibility of the *CES*; the *CES* cannot be reduced to a collection of its sub-structures without severely impeding the intrinsic causal flow. IIT prescribes a positive number $\Phi$ that measures the intrinsic irreducibility of the *CES*; higher values of $\Phi$ indicate higher levels of irreducibility.
- Again each experience is *definite* (an axiom), it has the set of phenomenal distinctions it has, neither less (a subset) nor more (a superset). Quantitatively, this means that the $\Phi$ value of the *CES,* corresponding to the particular experience, is greater than the $\Phi$ values of any other *CES* that has an overlap. This *maximally irreducible cause – effect structure* (**MICES**) of the *PSSC* is the winning structure and the value of its irreducibility, being maximal, is denoted by $\Phi^{max}$.
- IIT has a startlingly precise answer to the question of 'what consciousness is': the phenomenal consciousness is identical to the *MICES* of the corresponding *PSSC*.

- This identity should be understood in an explanatory sense: the intrinsic (subjective) feeling of the experience can be explained extrinsically (objectively, i.e., operationally or physically) in terms of cause–effect power (Albantaski 2023).
- Moreover, *the quantity or level of consciousness*—is measured by the $\Phi^{max}$ value of the *MICES*.
- Furthermore IIT provides a definite method for representing (unfolding) the *MICES* as a constellation of points bound by relations in the cause – effect space. This imparts the *MICES* a *form (shape)*. Now IIT plays its most intriguing card: The *quality* or *content of phenomenal consciousness* (*quale*) is specified by the *form* of the *MICES*: all phenomenal properties, such as the redness of red, the Eureka! feel of an insight, to mention a few, correspond to different forms of the *MICES*.
- Finally, the theory provides a calculus to evaluate the $\Phi^{max}$ value.

*C. Monitoring phenomenal agency using IIT*

Researchers in consciousness over the past decades have developed a plethora of well-developed theories of consciousness. These theories have different underpinnings (Seth & Bayne 2022). For example, in the Higher-order theories (HOT) that we encountered in the previous section, consciousness depends on meta-representations of lower-order mental states. In Global workspace theories (GWTs), consciousness depends on ignition and broadcast within a neuronal global workspace. In IIT, as I had noted before, consciousness is identical to the *MICES* of a physical substrate; IIT is elaborated, explained and argued for in the language of causal flow and this is the *appropriate* language for agency; consequently IIT develops a concept of consciousness that is anchored on agency.

Let us now go back to agentive phenomenology. From a problem solving perspective, the relevant phenomenal aspects of agency are *purposiveness* and *mineness* (see Section 2). To explain purposiveness, Mylopoulos (2020) cites David Hume ' …the internal impression we feel and are conscious of, when we knowingly give rise to any new motion of our body or new perception of our mind' (Hume 2000). Mylopoulos interprets that *giving rise to* is the key phrase that is fundamentally purposive. In IIT formalism, the 'giving rise to' is what the intrinsic causal flow achieves; past states of the *PSSC* give rise to future states. Hence the intrinsic causal flow is fundamentally purposive. Since the cause – effect structure determines the causal flow, that structure is fundamentally purposive. Thus the cause – effect structure or more appropriately the *MICES,* the one which is identified with phenomenal consciousness, captures the phenomenal aspects of *purposiveness.*

I explained (in section 2) the phenomenal characteristic of the sense of *mineness* as follows: there is something it is like to experiencing one's activity as *one's own actions* – the subjective feeling that I am the author of my own actions. In IIT parlance, my experience of my activities are captured by the intrinsic causal flows. The *MICES* determines all intrinsic causal flows, or it is the author of all such causal flows. I interpret this as the ownership of all causal flow. Hence *MICES* captures the fact that my own activities are experienced as my own. Thus *MICES* captures both purposiveness and mineness, it therefore captures the phenomenal aspects of agency.[4] Since conscious experience is identified with the *MICES*, it follows that agency is necessary for conscious experience in the IIT formalism. This is also the conclusion that Butlin et al. draw from their review: 'One argument for the claim that agency is necessary for consciousness is that this is implied by many scientific theories' (Butlin et al. 2003).

The measure of the level of consciousness $\Phi^{max}$, arises from the *MICES*, hence $\Phi^{max}$ is necessarily a measure of agency and, following the IIT formalism through, I conclude that the *quality* of the agency is specified within the *form* (shape) of the *MICES* when appropriately represented. In principle, these characteristics can be applied to any AI system to monitor the level and the nature of the phenomenal aspects of agency.

*Comments*: Before I end this section let me note that the formalism of IIT has been implicated with agency in ways that are very different to what I discussed above. For example Delafield-Butta & Trevarthenb (2022) advance the following point of view: 'But IIT adapts a linear causality, a "cause-effect" logic … Consciousness affords a further logic, an "action-effect" causality of experienced self-generated *agent action*, animated with feeling and vital purpose [my italics]'. What they mean is that the agent action generates 'action–effect' causality and not the 'cause-effect' causality of IIT. They are, of course, dealing with the third person aspects of agency and I have no disagreement with their proposition. They are silent on the cause-effect causality of IIT, which generates the first-person aspect of agency, its phenomenology, as I hold.

---

[4] Different authors add other characteristics to agency. See Mylopoulos (2020), for an extensive list. Adding more characteristics would mean a stronger sense of agency. However, as I discuss in the next section, just the character of purposiveness is enough for our objective.

Another view is expressed by Desmond & Huneman, who consider agency and consciousness as duals: 'If agency refers to the "activity" of the organism in relation to the environment, consciousness in its broadest sense denotes the "passivity" of the organism' (Desmond & Huneman 2022) They 'propose that measures of information integration (per IIT) can be more straightforwardly interpreted as measures of agency rather than of consciousness'. I disagree with this proposal; they are considering the third-person aspect of agency here and IIT holds no brief on this matter.

**4. Detecting whether or not an AI agent is altruistic or malicious**

There are obvious advantages and attendant risks to AI systems developing into superior problem solvers. Philosophers of artificial intelligence have speculated about the behaviours of hostile AI superintelligence. Presently, Bostrom's notion that AI superintelligence will take a 'treacherous turn' dominates (Bostrom 2014; Carlsmith 2022, Hendrycks et al. 2023). According to Bostrom (2014):

> (B)ehaving nicely while in the box is a convergent instrumental goal for friendly and unfriendly AIs alike. An unfriendly AI of sufficient intelligence realizes that its unfriendly final goals will be best realized if it behaves in a friendly manner initially, so that it will be let out of the box. It will only start behaving in a way that reveals its unfriendly nature when it no longer matters whether we find out; that is, when the AI is strong enough that human opposition is ineffectual.

From a behavioral standpoint, the malicious AI superintelligence imagined in this vignette is power-seeking, deceptive, manipulative and, above all, supplants any concern for humanity with its own (often anti-human) goals. On the other hand, there are no fundamental principles that prevent us from enabling the development of AI systems that are an aid to society. One would aspire to build altruistic AI systems that cherish human values. Counter to the 'treacherous turn', the Center for AI Safety proposes 'AI assistants [that] could act as advisors, giving us ideal advice and helping us make better decisions according to our own values. In general, AIs would improve social welfare and allow for corrections in cases of error or as human values naturally evolve' (Hendrycks et al. 2023). Contrasting the power-seeking behaviours of malicious AI agents, altruistic AI agents would behave selflessly, empathetically, cooperatively and, above all, put the needs of humanity above all else. It is therefore vital that we have the ability to monitor the agency, in particular, *purposiveness* of AI systems as they develop, to encourage the ones that cherish our humanness and regulate the ones that despise us.

How can we detect whether a putative AI system is altruistic or hostile? Behavioral analysis of their actions seems to be an obvious answer. To detect conscious agency, Butlin et al. have identifed behavioral indicators (Section 3) that involve; goal-oriented learning, responding flexibly to competing goals, and behavior that monitors and updates beliefs in response to the environment. But these agentive properties are hardly proprietary to either a hostile or an altruistic AI superintelligence. Similarly Carlsmith (2022) attributes two cognitive processes to hostile AI agents. Firstly, *strategic planning*. This refers to the capacity to 'make and execute plans, in pursuit of objectives, and on the basis of models of the world'. Secondly, *strategic awareness* refers to the capacity to model the 'causal upshot of gaining and maintaining power over humans and the real-world environment' (Carlsmith 2022). Again these are catch-all indicators that correctly target the strategic behaviours of a putative AI system as agentive, but do not distinguish those behaviours that are hostile, from those that are virtuous.

In principle, it is possible for researchers to hone the works cited above and come up with behavioral indicators that can differentiate between good and bad agents, as simply put. This will work if human monitors are not misled. In a recent discussion, Goldstein & Park (2023) argue that a range of existing AI systems have already learned how to deceive humans. The deceptive behavior of the present systems are easy to detect, however AI superintelligences can deceive their developers by initially feigning altruistic behaviours, to be 'let out of the box', so to speak. What we need are indicators that inform us, not only of how the agents behave but also of how they feel subjectively.

As discussed in Section 3, IIT directly addresses the phenomenal or subjective feeling of an experience and details it objectively through the $\Phi^{max}$ value and the *MICES* shape. $\Phi^{max}$ measures of the level of consciousness, and as I showed, it is necessarily also a measure of agency. This measure can help to specify, a scale for risk assessment and regulation, for future conscious AI systems. For example, through experimentation, it would be possible to fix a $\Phi^{max}$ value above which AI systems tend to gain the capability to use deception, to cheat on safety tests, say, so as to escape human control. At present the European Union is one of the few governmental organizations that specifies a regulatory framework through its AI Act; it assigns each AI system one of four risk levels: minimal, limited, high and unacceptable (European Commission 2023). One can similarly use the $\Phi^{max}$ measure to quantify risk levels, like the above, for future sentient AI systems.

The *quality* of the agency is specified by the *form* (shape) of the *MICES,* when appropriately represented. Each experience corresponds to a distinct shape. Whether an AI system develops to be altruistic or malicious towards humanity can in principle be inferred by fully unfolding the shape of *MICES* of that system. Phenomenally distinct experiences (e.g. agency) are represented by distinct sub-structures within the structure of the *MICES* (see Albantakis et al. 2023). The shapes, corresponding to different qualities of agency, are discovered through observation and manipulation, and these shapes provide the necessary *phenomenal indicators* to monitor the right kind of agency.

Two systems can be equivalent behaviorally, and yet be different architecturally. They will then have different *MICESes* and thus differ phenomenally. Phenomenal indicators, therefore, can detect the true nature of AI systems irrespective of their behavior. A conscious system cannot fake its phenomenology. We know firsthand that feigning altruism and acting altruistically are experiences that differ such that we ourselves do not confuse between them. Part of this phenomenological difference is that feigning altruism involves a *self-interested* purposiveness that is opposite to the *selfless* purposiveness associated with altruism proper. Phenomenal indicators would therefore be able to detect deceptive behavior.

Given an AI system, how to unfold its *MICES* and identify the phenomenal indicators is still a work in progress. An initial attempt in this direction shows how the feel of *spatial* experiences, in particular the feel that space is *extended*, can be accounted for by a corresponding feature of a *MICES* (Haun & Tononi 2019).[5] Although IIT provides a definite procedure to compute the $\Phi^{\max}$ value and compose the shape of the *MICES,* these computations become infeasible when one considers interesting systems that are not outright simple (Mayner et al. 2018). Ideally, one would like to conduct experimental studies to build a *reference library* that matches a particular kind of shape of the *MICES,* or a particular phenomenal indicator, with a particular kind of phenomenal feeling. 'Indeed, there is much scope for future research to begin mapping psycho- physics, … onto the geometry of shapes in cause – effect space….' (Tononi & Koch 2015). Given the complexity of the brain's neural architecture this is not yet computationally viable, except for a few initial but encouraging results (Massimini et al. 2010; Calasi et al. 2013). As AI systems develop to acquire consciousness, they will provide relatively simpler architecture to start experimenting and building such a library. With such a reference archive in place, developers can guide and nurture the evolution of an AI system such that it is steered towards achieving altruistic agency, while denying it any scope to achieve malicious agency.

---

[5] The following, as yet unpublished, papers are mentioned in the IIT literature that claim to account for the quality of experienced time [Comolatti et al.], and that of experienced objects [Grasso et al.].

# REFERENCES

Albantakis and others (2023) Integrated information theory (IIT) 4.0: Formulating the properties of phenomenal existence in physical terms. *arXiv preprint 2212.14787.* https://doi.org/10.48550/arXiv.2212.14787.

Altman S (2023) *Planning for AGI and Beyond.* Open AI. https://openai.com/blog/planning-for-agi-and-beyond (accessed 13 October 2023).

Aristodemou M (2014) *Law, Psychoanalysis, Society: Taking the unconscious seriously.* Routledge, Taylor & Francis Group.

Birch J, Ginsburg S and Jablonka E (2020) Unlimited Associative Learning and the origins of consciousness: a primer and some predictions. *Biology & Philosophy* 35(6).

Blackmore S and Troscianko ET (2018) *Consciousness: An Introduction.* (Third Edition) Routledge.

Bostrom N (2014) *Superintelligence: Paths, Dangers, Strategies.* Oxford University Press.

Bowman SR (2023) Eight things to know about Large Language Models. *arXiv preprint 2304.00612v1.* https://doi.org/10.48550/arXiv.2304.00612.

Bubeck S and others (2023), Sparks of Artificial General Intelligence: Early Experiments with GPT-4. *arXiv preprint 2303.12712v5.* https://doi.org/10.48550/arXiv.2303.12712.

Buderi R (1996) *The invention that changed the world: how a small group of radar pioneers won the Second World War and launched a technological revolution.* Simon & Schuster.

Butlin P and others (2023) Consciousness in Artificial Intelligence: Insights from the science of consciousness. *arXiv preprint 2308.08708v3.* https://doi.org/10.48550/arXiv.2308.08708.

Campbell M, Hoane AJ, and Hsu F (2002) Deep Blue. *Artificial Intelligence* 134(57).

Carlsmith J (2022) Is power-seeking AI an existential risk? *arXiv preprint 2206.13343v1.* https://doi.org/10.48550/arXiv.2206.13353.

Carruthers P and Gennaro R (2023) Higher-order theories of consciousness in Zalta EN and Nodelman U (Eds.), *The Stanford Encyclopedia of Philosophy.* https://plato.stanford.edu/archives/fall2023/entries/consciousness-higher/.

Carson SH (2014) Cognitive Disinhibition, Creativity, Psychopathology In Simonton DK (Ed), *The Wiley Handbook of Genius* (pp. 198-221). Wiley Blackwell.

Casali AG and others (2013) A theoretically based index of consciousness independent of sensory processing and behavior. *Science translational medicine* 5(198) pp. 198ra105.

Comolatti, R and others. Why does time feel flowing?; in preparation.

Chalmers DJ (1996) *The conscious mind: In search of a fundamental theory*. Oxford University Press.

Delafield-butt J and Trevarthen C (2022) Consciousness generates agent action. *The behavioral and brain sciences*, 45, e44. https://doi.org/10.1017/S0150525X2100203X.

Desmond H and Huneman P (2022) The integrated information theory of agency. *The Behavioral and brain sciences*, 22-24. https://doi.org/10.1017/S0140525X21002004.

European Commission (2023) *Regulation of the European parliament and of the council laying down harmonised rules on artificial intelligence (Artificial Intelligence Act) and amending certain union legislative acts*.

Future of Life Institute (2023, March 22nd) *Pause giant AI experiments: An open letter*. https://futureoflife.org/open-letter/pause-giant-ai-experiements. (Accessed 27th July 2023).

Goldstein EB (2018) *Cognitive psychology: Connecting mind, research and everyday experience* (5th edition) Cengage Learning.

Goldstein S and Park PS (2023, September 4th) *AI systems have learned to deceive humans. What does that mean for our future?* The Conversation. https://theconversation.com/ai-systems-have-learned-how-to-deceive-humans-what-does-that-mean-for-our-future-212197.

Grasso M and others. How do phenomenal objects bind general concepts with particular features?; in preparation.

Hanna A and Bender EM (2023, August 12th) *AI causes real harm. Let's focus on that over the end-of-humanity hype*. Scientific American. https://www.scientificamerican.com/article/we-need-to-focus-on-ais-real-harms-not-imaginary-existential-risks/ (Accessed 16th September 2023).

Haun A and Tononi G (2019) Why does space feel the way it does? Towards a principled account of spatial experience. *Entropy*, 21(12), 1160.

Hendrycks D, Mazeika M and Woodside T (2023) An overview of catastrophic AI risks. *arXiv preprint 2306.12001v6*. https://doi.org/10.48550/arXiv.2306.12001.

Horgan T (2011) From agentive phenomenology to cognitive phenomenology: a guide for the perplexed In Bayne T and Montague M (Eds.), *Cognitive phenomenology*. Oxford University Press.

Hume D (2000) *A Treatise of Human Nature*. Norton DF and Norton MJ (Eds.), Oxford University Press.

Hurley SL (1998) *Consciousness in action*. Harvard University Press.

Kounios J and Beeman M (2015) *The eureka factor: Aha moments, creative insight, and the brain.* Random House.

Malek MDA (2022) Criminal courts' Artificial Intelligence: the way it reinforces bias and discrimination. *AI and Ethics*, 233-245.

Massimini M and others (2010) Cortical reactivity and effective connectivity during REM sleep in humans. *Cognitive neuroscience* 1(3), 176-183. https://doi.org/10.1080/17588921003731578.

Mayner WGP and others (2018) PyPhi: a toolbox for integrated information theory. *PLOS Computational Biology* 14(7). https://doi.org/10.1371/journal.pcbi.1006343.

Mednick S (1962) The associative bias of the creative process. *Psychological Review*, 69, 220.

Mylopoulos M and Shepherd J (2020) The experience of agency In Kriegel U (Ed.) *The Oxford Handbook of the Philosophy of Consciousness.* Oxford University Press. 164-187.

Ohlsson S (2011) *Deep Learning: How the mind overrides experience.* Cambridge University Press.

OpenAI (2023) GPT-4 Technical Report. *arXiv preprint 2303.08774.* https:///doi.org/10.48550/arXiv.2303.08774.

Seth AK and Bayne T (2022) Theories of consciousness. *Nature reviews neuroscience.* 439-452.

Stokes D (2014) The role of imagination in creativity In Paul ES and Kaufman BS (Eds.) *The philosophy of creativity: new essays.* Oxford University Press.

Tononi G and Koch C (2015) Consciousness: here, there, and everywhere? *Phil. Transc. R. Soc. B.* http://doi.org/10.1098/rstb.2014.0167.

Van Gelder T (1999) Wooden iron? Husserlian phenomenology meets cognitive science. In Petitot J and others (Eds.) *Naturalizing phenomenology: issues in contemporary phenomenology and cognitive science.* Stanford University Press.

Van Gulick R (2007) 30. Functionalism and Qualia. In Velmans M and Schneider S (Eds.) *The Blackwell Companion to Consciousness.* Wiley Blackwell.

Wegbreit E and others (2012) Visual attention modulates insight versus analytic solving of verbal problems. *The Journal of Problem Solving.*

Weisberg RW (2018) Expertise and structured imagination in creative thinking: reconsideration of an old question. In Ericsson KA and others (Eds.) *The Cambridge Handbook of Expertise and Expert Performance* (Second Edition). Cambridge University Press.